\documentclass{clv3}

\usepackage{hyperref}
\usepackage{xcolor}

\definecolor{darkblue}{rgb}{0, 0, 0.5}
\hypersetup{colorlinks=true,citecolor=darkblue, linkcolor=darkblue, urlcolor=darkblue}

\usepackage{algorithmic}

\dochead{Last Words}

\runningtitle{Revisiting the ASR/NLU Boundary }

\runningauthor{Faruqui and Hakkani-T\"ur}

\begin{document}
\title{Revisiting the Boundary between ASR and NLU in the Age of Conversational Dialog Systems}

\author{Manaal Faruqui}
\affil{Google Assistant\\\texttt{mfaruqui@google.com}}

\author{Dilek Hakkani-T\"ur}
\affil{Amazon Alexa AI\\\texttt{hakkanit@amazon.com}}


\maketitle
\begin{abstract}
As more users across the world are interacting with dialog agents in 
their daily life, there is a need for better speech understanding that
calls for renewed attention to the dynamics 
between research in automatic speech recognition (ASR) and natural 
language understanding (NLU). We briefly review these research areas
and lay out the current relationship between them. In light of the 
observations we make in this paper, we argue that (1) NLU should be 
cognizant of the presence of ASR models being used upstream in a 
dialog system's pipeline, (2) ASR should be able to learn from errors
found in NLU, (3) there is a need for end-to-end datasets that provide
semantic annotations on spoken input, (4) there should be stronger
collaboration between ASR and NLU research communities.
\end{abstract}

\section{Introduction}
More and more users every day are communicating with conversational dialog
systems present around them like Apple Siri, Amazon Alexa, and Google Assistant.
As of 2019, 31\% of the broadband households in the US have a digital assistant.\footnote{ https://www.statista.com/statistics/791575/us-smart-speaker-household-ownership/}
Henceforth, we refer to these systems as \textit{dialog agents} or simply \textit{agents}.
A majority of queries issued to these
dialog agents are in the form of speech as the users are directly talking to these
agents hands-free.

This is in contrast to a few years ago when most of the traffic
to search engines like Google Search, Yahoo, or Microsoft Bing was in the form of
text queries. The NLU models that underlie these search engines 
were tuned to handle textual queries typed by users. However, with
the changing nature of query stream from text to speech,
these NLU models also need to adapt in order to better understand the user.

\begin{figure}[!t]
    \centering
    \includegraphics[scale=0.3]{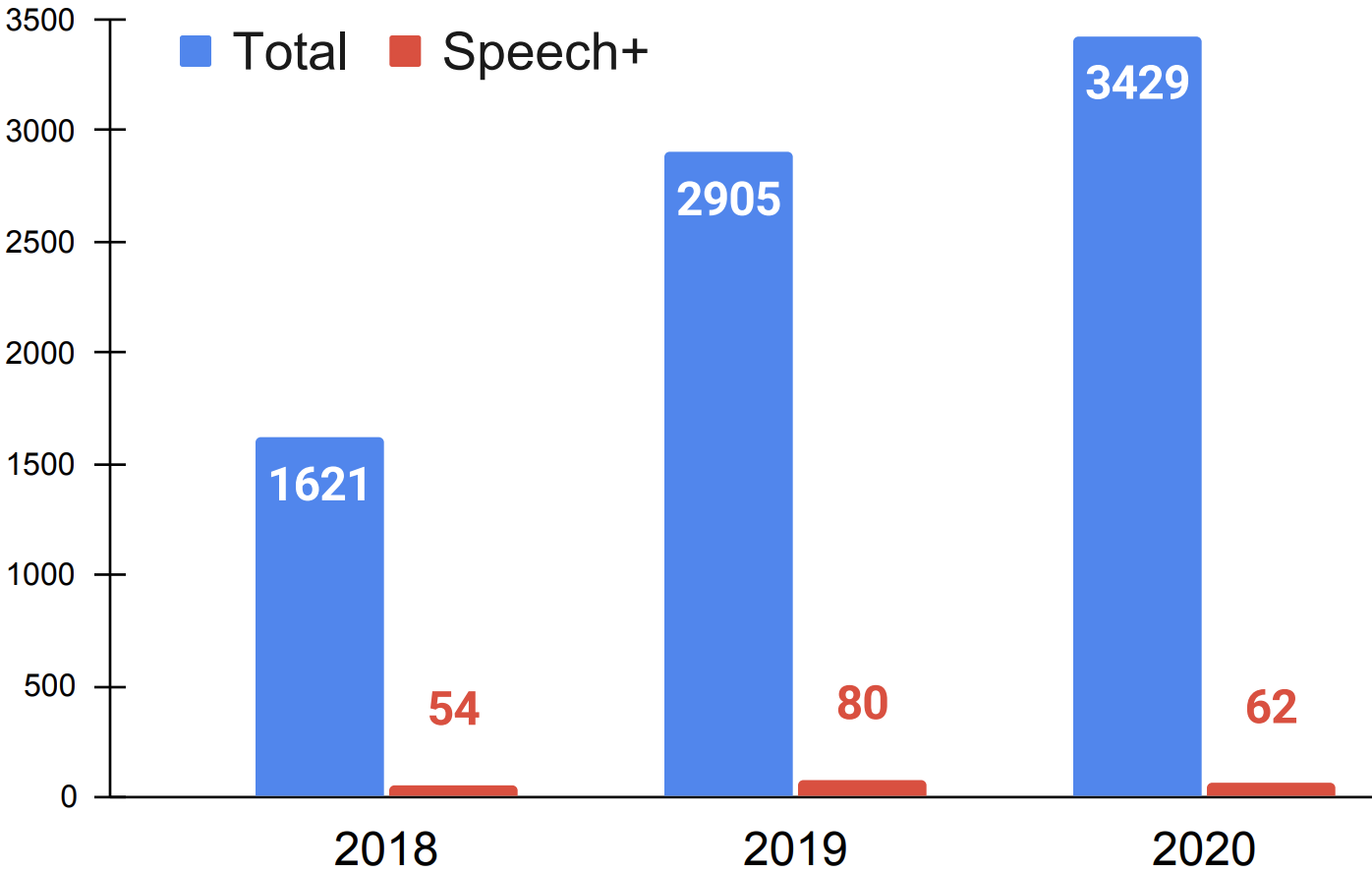}
    \caption{The number of submitted papers in the speech processing (+ multimodal) track vs. total in ACL conference from 2018-2020.}
    \label{fig:speech_in_nlp}
\end{figure}

This is an opportune moment to bring our attention to the current state of 
ASR and NLU, and the interface 
between them. Traditionally, the ASR and NLU research communities have operated pretty
independently with little cross-pollination. 
While there is a long history of efforts to get ASR and NLU researchers to collaborate, 
for example, through conferences like HLT and DARPA programs \cite{liu2006enriching,ostendorf2008speech}, the two communities are diverging again. 
This is reflected in their disjoint set of 
conference publication venues: ICASSP, ASRU/SLT, Interspeech are the major conferences for
speech processing, whereas ACL, NAACL, and EMNLP being the major venues for NLU.
Figure~\ref{fig:speech_in_nlp} shows the total number of submitted papers to the ACL
conference, and in the speech processing track from 2018-2020 at the same conference.\footnote{The speech processing area also included papers from other related areas like 
multimodal processing, so the numbers presented here are an upper bound on speech processing papers.}
The number of submitted speech related papers in the conference constitute only
$54$ ($3.3\%$), $80$ ($2.7\%$), and $62$ ($1.8\%$) in 2018, 2019 and 2020 respectively, 
showing the limited amount of interaction between these fields.

In this paper, we analyze the current state of ASR, NLU, and the relationship between
these two large research areas. We classify the different paradigms in which ASR and NLU
operate currently and present some ideas on how these two fields can benefit from 
transfer of signals across their boundaries. We argue that a closer collaboration and 
re-imagining of the boundary between speech and language processing is critical for the 
development of next generation dialog agents, and for the advancement of research in
these areas. 

Our call is specially aimed at the computational linguistics community to consider peculiarities of spoken language, such as, disfluencies and prosody that may carry additional information for NLU, and errors associated with speech recognition as core part of the language understanding problem. This change of perspective can lead to the creation of datasets that span across the ASR/NLU boundary, which in turn will bring the NLU systems closer to real-world setting as well as increase collaboration between industry and academia.

\section{Changing Nature of Queries}

As users move from typing their queries to conversing with their agent through dialog, there
are a few subtle phenomena which differ in the nature of these queries. We briefly discuss these in the following section.

\begin{table*}[!t]
  \centering
  \begin{tabular}{c|c}
  \hline
  Typed & Spoken\\
  \hline
  \hline
  barack obama age & what is the age of barack obama\\
  \hline
  boston denver flight & book a flight \textit{from london to umm no} from boston to denver\\
  \hline
  scooby doo breed & tell me what's the breed of scooby doo\\
  \hline
  \end{tabular}
  \caption{Examples of typed and spoken queries that have the same user intent. Spoken queries have phenomenon like disfluencies which are
  not part of typed queries.}
  \label{tab:typed-spoken}
\end{table*}

\subsection{Structure of the Query}
User-typed queries aren’t always well-formed nor do they
always follow the syntactic and semantic rules of a language \cite{bergsma2007learning,barr-jones-regelson:2008:EMNLP,manshadi2009semantic,mishra2011unsupervised}. This is not surprising
because users often want to invest as little effort as they can in typing a query. They
provide the minimum required amount of information in the form of words that they think can
return the desired result, and hence, typed search queries are mostly bag of keywords
\cite{baeza2006intention,ZENZ2009166}.

On the other hand, spoken utterances addressed to dialog agents are closer to natural language, 
contain well-formed sequence of words that form grammatical natural language sentences (though spoken language can also be ungrammatical cf. \S\ref{sec:errors}), and
are more complex and interactive \cite{trippas2017how,trippas2018informing}. For example,
sometimes a user utterance might need to be segmented into multiple parts:
\begin{center}
    Agent: \textit{``What movie genre do you like?''}\\
    User: \textit{``I don't know <pause> anime''}
\end{center}
Here, if we don't explicitly use the <pause> marker to segment the user utterance, we might
understand the user intent as \textit{``I don't know anime''} instead of the selected genre
being \textit{``anime''}. Thus,
traditional techniques of information retrieval are not effective in understanding
such conversational utterances which need deeper understanding.
Table~\ref{tab:typed-spoken} shows how the same intent is surfaced in typed 
vs. spoken utterances.

Another common phenomenon in spoken utterances is disfluencies. When a user stammers, repeats, corrects themselves or changes their mind in between the utterance, they introduce disfluency in the utterance \cite{Schriberg1994PreliminariesTA}. For example, as shown in Table~\ref{tab:typed-spoken} \textit{``book a flight from london to umm no from boston to denver''} is a disfluent query. There has been limited NLU
research done on correcting speech recognition errors or handling disfluencies.
For example, in the last 27 years there has been only one major dataset available containing annotated
disfluencies in user utterances \cite{switchboard,Schriberg1994PreliminariesTA}.

\subsection{Errors in the Query}
\label{sec:errors}
Not only do spoken vs. typed queries differ in structure and style, 
they also vary in the kind of
noise or anomalies in the input. While typed queries can contain spelling errors, spoken
queries can contain \textbf{speech recognition errors}, and \textbf{endpointing} issues, which we discuss below.

While there has been
extensive research on correcting spelling errors \cite{hladek2020survey} including state-of-the-art neural machine
translation models being launched in products \cite{lichtarge-etal-2019-corpora},\footnote{https://cloud.google.com/blog/products/productivity-collaboration/using-neural-machine-translation-to-correct-grammatical-in-google-docs} there has been limited NLU
research done on correcting speech recognition errors.

Handling speech recognition errors is crucial for downstream NLU models to work 
effectively because an error in the transcription of a single word can entirely 
change the meaning of a query. For example, a user utterance:
\textit{``stair light on''} is transcribed by the cloud Google Speech 
Recognizer\footnote{https://cloud.google.com/speech-to-text} as:
\textit{``sterilite on''}. In this example, if the NLU model is given the 
query \textit{``sterilite on''}, it is very
hard for it to uncover the original intent of the user which was to 
turn on the stair lights, unless we force the NLU model to learn to 
correct/handle such systematic ASR errors. Such error analysis is often done
for industrial query logs \cite{Shokouhi2014MobileQR} but these datasets are
not publicly available for academic research purposes.

Another common error that affects NLU is speech endpointing. If a user 
utterance is pre-planned (such as, \textit{``play music''}), the users do not hesitate, 
but if the utterance is complex or the user is responding to a request from 
the agents,such as, \textit{``do you prefer 5 or 7?''}), the users may hesitate when responding 
and pause, saying \textit{``oh <long pause> 7''} causing the ASR model to assume
that the user stopped after saying \textit{``oh''} which lead to incomplete 
queries being transmitted to NLU. On the other hand, if we can learn to provide 
a signal to endpointing from the dialog manager that the recognized utterance is missing a value (in 
this case, probably an integer) according to the conversation context, we can 
improve endpointing, and hence recognition and downstream NLU. Datasets pertaining 
to endpointing phenomenon are currently available in very limited domains 
\cite{raux-eskenazi-2008-optimizing} and there is a need for such experimentation 
in broader domains and the impact analysis of such errors on overall user experience.

Thus, it is imperative for NLU models to be aware of the fact that 
there could be speech recognition errors in the input that might need
to be corrected.

\begin{figure*}[!tb]
    \centering
    \includegraphics[scale = 0.4]{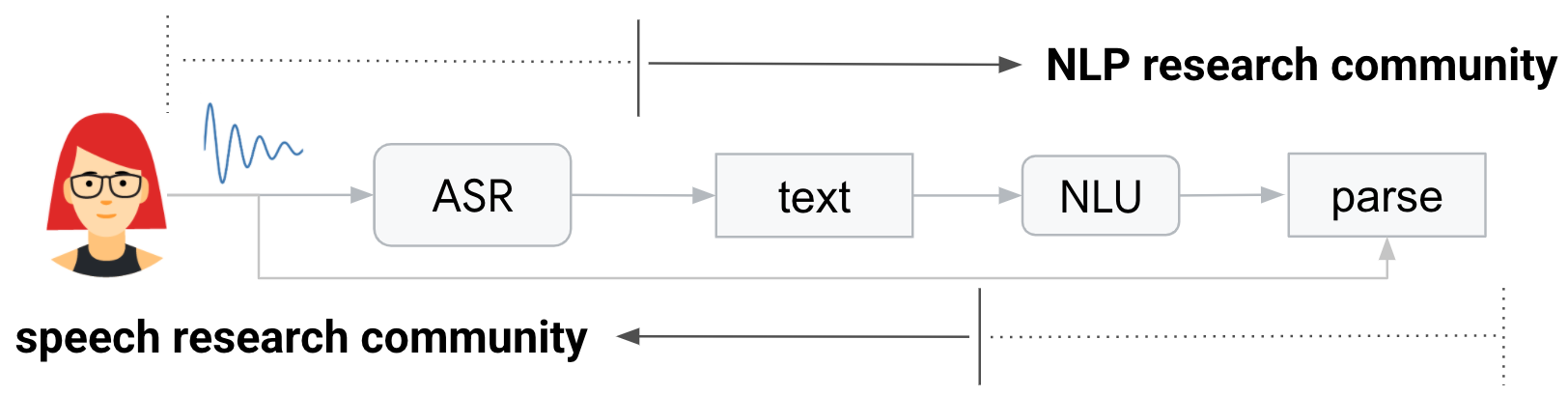}
    \caption{The current focus of speech and NLU research community (dark lines) and
    preferred focus of speech and NLU community (dotted lines) in future.}
    \label{fig:schema}
\end{figure*}

\section{Current State of Research}
\label{sec:current-research}

We now describe the current core areas of research related to
understanding the meaning of a written piece of text or spoken
utterance. Figure~\ref{fig:schema} shows a general spoken language
understanding pipeline.

\subsection{Natural Language Understanding}
\label{ref:nlu}
Natural language understanding refers to the process of understanding the 
semantic meaning of a piece of text. More concretely, understanding the semantic
meaning often implies semantic parsing in NLP academic research and industry. 
Semantic parsing
is the task of converting a natural language utterance to a logical form: a 
machine-understandable representation of its meaning. Semantic parsing
is a heavily studied research area in NLP \cite{kamath2018survey}.

In its simplest form, a semantic parse of a given query can contain a
label identifying the desired \textit{intent} of the query and the
\textit{arguments} that are required to fulfill the given intent. However,
a more informative semantic parse can contain edges describing relations
between the arguments, as in the case of abstract meaning representations (AMR)
\cite{banarescu-etal-2013}, or a graph describing how different words join together to construct
the meaning of a sentence, as in the case of combinatory categorial grammar (CCG)
\cite{Steedman1987CombinatoryGA}.\footnote{Reviewing all different semantic parsing formulations is beyond the scope
of this paper.} Figure~\ref{fig:nlu-slu} shows a parse containing a semantic role labels
of a given query obtained using the AllenNLP tool \cite{Gardner2017AllenNLP}. 

In general, all the semantic parsing research done in NLU assumes the 
input to the parser as a piece of text (and available world context). Thus, the training/evaluation dataset of all
available semantic parsing datasets are devoid of any speech phenomenon like
the presence of speech recognition errors, annotation of pauses or disfluencies.

\subsection{Spoken Language Understanding}
\label{ref:slu}

\begin{figure*}[!tb]
    \centering
    \includegraphics[scale = 0.65]{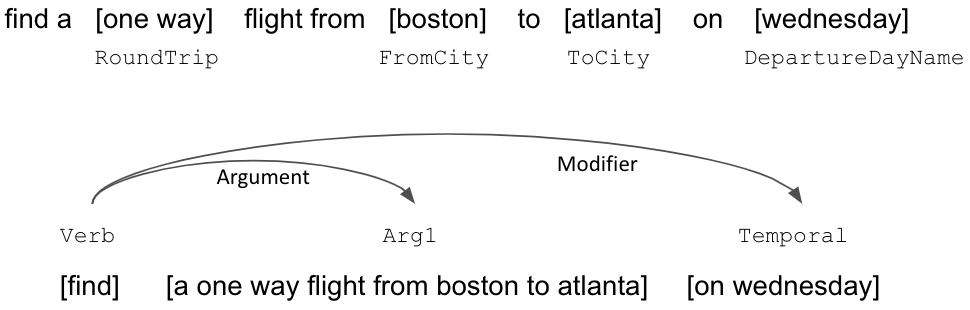}
    \caption{SLU annotation (top) and NLU semantic role labeling annotation (bottom) on a sentence from the English ATIS corpus \cite{price-1990-evaluation}, a popular SLU benchmark.}
    \label{fig:nlu-slu}
\end{figure*}

Spoken language understanding refers to the process of identifying the meaning
behind a spoken utterance \cite{slu,slu-book}. In that regard, the end goal of
NLU and SLU is the same but the input to NLU and SLU components are different:
text in the former, and speech input in the latter. The most common SLU task
is intent prediction and slot-filling which involves classifying the intent of
the utterance and identifying any required arguments to fulfill that intent
\cite{price-1990-evaluation}. Figure~\ref{fig:nlu-slu} shows the SLU annotation
for a slot-filling task. We now review the main approaches used to solve SLU. 

\subsubsection{Speech $\rightarrow$ Text $\rightarrow$ Semantic Parse}
The traditional way of performing speech understanding is to use a 
pipeline approach: first use an ASR system to transcribe the speech 
into text and then run NLU on the transcribed text to result into a
semantic parse. 
Using a pipeline approach has its own set of pros and 
cons. Using a 2-step pipeline approach is modular. The first
component is an ASR system and the second component is an NLU system.
The errors of each module can be independently analyzed and corrected
\cite{fazel2019investigation,wang-etal-2020-data}.
The training of both models is also independent, which makes it easier to
use off-the-shelf start-of-the-art ASR and NLU models during inference.

The obvious disadvantage of this method is that the NLU and ASR models
are unaware of each other. Since the models are not trained jointly, the ASR model
cannot learn from the fact that the downstream NLU model could have failed
on an erroneous ASR prediction. Similarly, at inference time the NLU model
relies only on the best prediction of the ASR model and cannot exploit the
uncertainty in ASR's prediction. However, this can be fixed to a good extent
by forcing the ASR model to propagate the $n$-best list of speech transcript
hypotheses to the NLU system and let the NLU model use all the hypotheses 
together to make the semantic parse prediction 
\cite{hakkani2006beyond,deoras2012joint,9053213,li2020improving}
or using a lattice or word -confusion network as input \cite{tur2013semantic,Ladhak+2016}.

\subsubsection{Speech $\rightarrow$ Semantic Parse}
There is a renewed focus of attention on approaches to directly parse the
spoken utterance to derive the semantics by making a deep neural network
consume the speech input and output the semantic parse
\cite{8639043,chen2018spoken,8461785,Kuo2020}. This is an end-to-end
approach that does not rely on the intermediate textual representation
of the spoken utterance produced by an ASR system. Thus, this system
can be trained end-to-end with direct loss being optimized on the
semantic parse prediction. However, such end-to-end models are data-hungry 
and suffer from lack of training data \cite{lugosch2019speech,Li2020OnTC}.
Even though such models often have better performance on benchmark 
datasets, deploying such models in a user-facing product is hard because
of the lack of ease of debugging and fixing errors in output 
\cite{Glasmachers2017LimitsOE}. 

\section{Reimagining the ASR-NLU Boundary}

In section \ref{sec:current-research} we saw that parsing the user query is a step
in SLU. And thus SLU is a large umbrella which utilizes NLU techniques to 
parse spoken language. Even though both SLU and NLU at some level are
solving the same problem, there is a clear disconnect between the way
problems are formulated and devised solutions for these problems.
On one hand, in NLU, a lot of emphasis is laid on understanding deep
semantic structures in text formulated in the tasks of semantic
parsing, dependency parsing, language inference, question answering,
coreference resolution etc. On the other hand, SLU mainly concerns with
information extraction on the spoken input formulated in the tasks of
slot filling, dialog state modeling etc.

Even though there are academic benchmarks available for spoken language
understanding which aim to extract information from the spoken input,
there is an informal understanding between the ASR and NLU communities
that assumes that as long as the ASR component can transcribe the spoken
text correctly, majority of the language understanding burden can be 
taken up by the NLU community.
Similarly, there is an implicit assumption in the NLU community that
ASR will provide correct transcription of the spoken input and hence
NLU does not need to account for the fact that there can be errors in
the ASR prediction. We consider the absence of an explicit two-way 
communication between the two communities problematic.

Figure \ref{fig:schema} shows how the NLU research community can 
expand its domain to also consider spoken language as input to the
NLU models instead of pure text. Similarly, the ASR community can 
also account for whether the text produced by them can result in 
a semantically coherent piece of text or not by explicitly trying
to parse the output. That said, there are already a few efforts
which have tried blurring the boundary between ASR and NLU. We will
first give some examples of how ASR and NLU can learn from each
other and then review some existing initial work in this domain
aimed at enriching existing academic benchmark datasets.

\subsection{ASR to NLU Transfer}
A significantly large missing portion of information in understanding
a spoken input is the nature of speech, which we often refer to as 
\textit{prosody}. Whether the user was angry, happy, in a rush,
frustrated etc. can help us better understand what the user's real
intent was. For example, \textit{``no... don't stop''} and \textit{``no don't... stop''} have exactly opposite meanings depending on whether the user
paused between first and second words or second and third words. This information can only be transferred
from speech to NLU. Amazon Alexa already has such a tone-detection feature
deployed in production.\footnote{https://onezero.medium.com/heres-how-amazon-alexa-will-recognize-when-you-re-frustrated-a9e31751daf7} There are 
academic datasets that map speech to emotion 
\cite{livingstone2018ryerson}, however, academic benchmarks containing examples of intonation affecting the NLU output do not exist.

An ASR system can provide more information than its best guess to the
NLU model by providing a list of $n$-best speech hypotheses. Unfortunately,
most of the state-of-the-art NLU models are trained to only accept a
single string of text as input, be it parsing, machine translation, or
any other established NLU task. To some extent, SLU has enlarged the domain
for understanding tasks by creating benchmark datasets that contain 
$n$-best speech hypotheses lists, for example, the dialog state tracking 
challenge dataset DSTC-2 \cite{henderson2014second}. This allows the
language understanding model to make use of all the $n$-best instead of 
just relying on the top ASR output.\footnote{Note that lack of diversity in the n-best speech hypotheses could be an issue and directly using the recognition lattice produced by ASR might be more informative.}

\subsection{NLU to ASR Transfer}
Making sure that the output produced by ASR model can be
understood by an NLU model can help improve transcription quality 
\cite{Velikovich2018SemanticLP}. For example, trying to boost paths
in the ASR lattice that contain named entities as predicted by a
named entity recognition (NER) model can help overcome recognition
errors related to out-of-vocabulary words \cite{Serrino2019ContextualRO}.

ASR models can also learn from the errors produced in the NLU model. If
a downstream NLU model in a conversational agent cannot reliably parse
a given ASR output, this might indicate the presence of
speech recognition errors. If there is a reliable way to identify 
the cause for NLU failure as a speech error, then such examples can be
provided back to the ASR module to improve itself. 
In Google Home, \citet{faruqui2020contextual} propose a method for
picking out the correct transcription from the $n$-best hypotheses if
the top hypothesis does not parse, and explicitly confirm the new hypothesis
with the user in a dialog. If the user accepts the selected speech hypothesis,
this correction is provided as a training example to the ASR system.
In general, if the ASR models start computing error based on whether
or not the produced text can be semantically parsed, the performance
metric will be more representative of the real-world setting instead
of the currently widely used word-error-rate (WER) metric 
\cite{He2011WhyWE}.

\subsection{Spoken NLU Datasets}

There has already been some progress on the front of enriching existing NLU
benchmarks with speech. We would now briefly review these efforts.

\subsubsection{Datasets with Speech Input}

\paragraph {Synthesized Speech} \citet{Li2018SpokenSA} presented a new dataset called \textit{Spoken-SQuAD},
which takes the existing NLU dataset SQuAD \cite{rajpurkar2016squad} containing
textual questions and textual document. The Spoken-SQuAD dataset
contains audio form of the document that has been artificially constructed
by using Google text-to-speech system, and then the textual form of the document
was generated using CMU Sphinx speech recognizer \cite{walker2004sphinx}. 
\citet{You2020TowardsDD} have created
Spoken-CoQA dataset from the CoQA dataset \cite{reddy-etal-2019-coqa} 
using the same technique. Both of these systems have shown that the presence 
of ASR errors has devastating effect on the quality of the QA system. However, it is worth
noting that this speech still does not reflect what people do in spontaneous interactions.

\paragraph {Natural Speech} The above datasets contain artifically synthesized speech. The
OSDQA dataset \cite{lee2018odsqa} on the other hand was constructed by recruiting 20
speakers for speaking out the documents from the original QA dataset \cite{shao2018drcd}.
This dataset is for Chinese QA and contains spoken Chinese documents as audio.
In order to accurately model real-world setting of SLU, we need to construct
dataset containing real spoken utterances similar to the approach used in OSDQA.

There are certain drawbacks with both the \textit{artificial} and \textit{natural}-speech style 
datasets. While artificially generated speech suffers from lack of sufficient speech style
variability and the absence of natural speech cues, naturally generated speech comes with 
strict privacy and scalability concerns preventing a large-scale collection of human speech. 
This privacy concern is even more pressing when dealing with utterances that humans issue to 
dialog agents at home that contain personal information.

\subsubsection{Datasets with Speech Errors} 
Instead of providing audio input in the dataset, another line of effort is about 
adding speech recognition errors in the transcribed text. For example, Raddle
\cite{peng2020etal} is a benchmark dataset and an evaluation platform for dialog 
modeling where the
input text can contain speech phenomena like verbosity, and speech recognition errors. 
Similarly, the LAUG toolkit \cite{Liu2020RobustnessTO} provides options to evaluate dialog
systems against noise perturbations, speech characteristics like repetitions, corrections 
and language variety. NoiseQA \cite{ravichander-etal-2021-noiseqa} contains ASR errors in the questions of the QA dataset introduced both using synthetic speech and natural speech.

\section{Conclusion}

In this paper we have argued that there is a need for revisiting
the boundary between ASR and NLU systems in the research community. We are
calling for stronger collaboration between the ASR and NLU communities
given the advent of spoken dialog agent systems that need to understand
spoken content. In particular, we are calling for NLU benchmark datasets
to revisit the assumption of starting from text, and instead move towards
a more end-to-end setting where the input to the models is in the form
of speech as is the case in real-world dialog settings.

\acknowledgments

We thank the anonymous reviews for their helpful feedback. We thank Shachi Paul, Shyam Upadhyay, Amarnag Subramanya, Johan Schalkwyk, and Dave Orr for their comments on the initial draft of the paper.

\bibliography{custom}

\end{document}